\documentclass[sigconf, nonacm, natbib]{acmart}

\AtBeginDocument{%
  }

\usepackage{microtype}
\usepackage{algorithmic}
\usepackage{algorithm}
\usepackage{amsmath}
\usepackage{booktabs}

\usepackage{amssymb}

\begin{document}

\title{OrthoGeoLoRA: Geometric Parameter-Efficient Fine-Tuning for Structured Social Science Concept Retrieval on the Web}

\author{Zeqiang Wang}
\affiliation{%
  \institution{University of Surrey}
  \country{UK}
}

\author{Xinyue Wu}
\affiliation{%
  \institution{Washington State University}
  \country{USA}
}

\author{Chenxi Li}
\affiliation{%
  \institution{University of Oxford}
  \country{UK}
}

\author{Zixi Chen}
\affiliation{%
  \institution{New York University Shanghai}
  \country{China}
}

\author{Nishanth Sastry}
\affiliation{%
  \institution{University of Surrey}
  \country{UK}
}

\author{Jon Johnson}
\affiliation{%
  \institution{University College London}
  \country{UK}
}

\author{Suparna De}
\affiliation{%
  \institution{University of Surrey}
  \country{UK}
}


% \renewcommand{\shortauthors}{Wang et al.}
\renewcommand\footnotetextcopyrightpermission[1]{}

\begin{abstract}

Large language models and text encoders increasingly power web-based information systems in the social sciences, including digital libraries, data catalogues, and search interfaces used by researchers, policymakers, and civil society. Full fine-tuning is often computationally and energy intensive, which can be prohibitive for smaller institutions and non-profit organizations in the Web4Good ecosystem. Parameter-Efficient Fine-Tuning (PEFT), especially Low-Rank Adaptation (LoRA), reduces this cost by updating only a small number of parameters. We show that the standard LoRA update $\Delta W = BA^\top$ has geometric drawbacks: gauge freedom, scale ambiguity, and a tendency toward rank collapse. We introduce OrthoGeoLoRA, which enforces an SVD-like form $\Delta W = B\Sigma A^\top$ by constraining the low-rank factors to be orthogonal (Stiefel manifold). A geometric reparameterization implements this constraint while remaining compatible with standard optimizers such as Adam and existing fine-tuning pipelines. We also propose a benchmark for hierarchical concept retrieval over the European Language Social Science Thesaurus (ELSST), widely used to organize social science resources in digital repositories. Experiments with a multilingual sentence encoder show that OrthoGeoLoRA outperforms standard LoRA and several strong PEFT variants on ranking metrics under the same low-rank budget, offering a more compute- and parameter-efficient path to adapt foundation models in resource-constrained settings.

\end{abstract}

\begin{CCSXML}
<ccs2012>
   <concept>
       <concept_id>10010147.10010178.10010179</concept_id>
       <concept_desc>Computing methodologies~Natural language processing</concept_desc>
       <concept_significance>500</concept_significance>
       </concept>
   <concept>
       <concept_id>10002951.10003317</concept_id>
       <concept_desc>Information systems~Information retrieval</concept_desc>
       <concept_significance>500</concept_significance>
       </concept>
   <concept>
       <concept_id>10002951.10003260.10003261</concept_id>
       <concept_desc>Information systems~Web searching and information discovery</concept_desc>
       <concept_significance>500</concept_significance>
       </concept>
   <concept>
       <concept_id>10003456</concept_id>
       <concept_desc>Social and professional topics</concept_desc>
       <concept_significance>500</concept_significance>
       </concept>
 </ccs2012>
\end{CCSXML}

\ccsdesc[500]{Computing methodologies~Natural language processing}
\ccsdesc[500]{Information systems~Information retrieval}
\ccsdesc[500]{Information systems~Web searching and information discovery}
\ccsdesc[500]{Social and professional topics}

\keywords{Parameter-Efficient Fine-Tuning, Geometric Deep Learning, Information Retrieval, Computational Social Science, Structured Representation Learning}


\maketitle

\section{Introduction}

Large-scale pre-trained language models and text encoders have become core infrastructure for the modern web, supporting search, recommendation, and semantic retrieval across digital catalogues and social platforms~\cite{devlin-etal-2019-bert,radford2019language,NEURIPS2020_1457c0d6}. In the social sciences in particular, web-based catalogues and data portals increasingly rely on such models to help researchers, policymakers, and NGOs navigate large collections of reports, survey data, and grey literature. Yet the computational and energy cost of full fine-tuning remains prohibitive for many actors in this ecosystem, especially small institutions and non-profit organizations that are central to a web that serves the public good.

Parameter-Efficient Fine-Tuning (PEFT) methods address this challenge by updating only a small fraction of a model's parameters while aiming to match the performance of full fine-tuning. Among these techniques, Low-Rank Adaptation (LoRA)~\cite{DBLP:journals/corr/abs-2106-09685} has emerged as a simple and widely adopted approach: it freezes the pre-trained weights $W_0$ and injects a trainable low-rank update $\Delta W = BA^\top$, where the rank $r$ is much smaller than the matrix dimensions. LoRA has quickly become the default PEFT strategy in many web and GenAI applications. However, despite its practical success, the geometric properties of its parameterization and their implications for optimization and representation quality remain underexplored.

In this work, we take a geometric perspective on LoRA and argue that the standard parameterization $\Delta W = BA^\top$ exhibits three structural issues that are particularly harmful when learning structured knowledge: (i) \emph{gauge freedom}, where many different pairs $(A,B)$ represent the same update, creating flat valleys in the loss landscape; (ii) \emph{scale ambiguity}, where direction and magnitude are confounded because scaling $B$ up while scaling $A$ down yields the same $\Delta W$; and (iii) a tendency towards \emph{rank collapse}, where unconstrained gradient descent encourages columns of $A$ and $B$ to become collinear, reducing the effective rank of the update and wasting the configured capacity.

To ground our analysis in a concrete Web4Good scenario, we focus on \emph{hierarchical concept retrieval} over the European Language Social Science Thesaurus (ELSST), a curated ontology used to index social science resources in web-based repositories. Given a nuanced textual description, the model must retrieve the correct ELSST concept. This setup goes beyond simple keyword matching: it requires encoding subtle, ontology-aware relationships that reflect how social scientists organize knowledge about topics such as inequality, migration, or public health. High-quality retrieval in this setting directly benefits open scientific infrastructures and evidence-informed policy work, aligning with Web4Good's emphasis on web technologies for societal impact and computational social science.

Motivated by the optimal structure of low-rank approximations, we propose \textbf{OrthoGeoLoRA}, a principled reformulation of low-rank adaptation inspired by the Eckart--Young--Mirsky theorem. Instead of the unconstrained $\Delta W = BA^\top$, OrthoGeoLoRA parameterizes the update in an SVD-like form $\Delta W = B \Sigma A^\top$, where $A$ and $B$ are constrained to lie on Stiefel manifolds (orthonormal columns) and $\Sigma$ is a learnable diagonal matrix of singular values. We implement these constraints through geometric reparameterization: unconstrained Euclidean parameters are mapped to orthonormal factors at each forward pass via differentiable orthogonalization, so that standard optimizers such as Adam can be used without modification and with nearly identical training pipelines and parameter counts as vanilla LoRA.

Our contributions are three-fold:
\begin{itemize}
    \item \textbf{Geometric insight into LoRA.} We provide a rigorous analysis of the geometric deficiencies inherent in the standard LoRA formulation---gauge freedom, scale ambiguity, and rank collapse---and show how an SVD-informed structure addresses them. This clarifies \emph{why} certain LoRA variants help in practice and identifies which issues remain unsolved.
    \item \textbf{A principled PEFT method for structured concept representations.} We introduce OrthoGeoLoRA, which enforces orthogonality on low-rank factors via Stiefel-manifold reparameterization while remaining a drop-in replacement for LoRA in existing PEFT pipelines. This design eliminates redundant parameterizations, decouples direction from magnitude, and prevents rank collapse by construction.
    \item \textbf{A socially grounded evaluation and empirical validation.} We construct a hierarchical concept retrieval benchmark based on ELSST, reflecting a realistic social science web application, and perform an expert-based validation of the synthetic descriptions. On this benchmark, OrthoGeoLoRA consistently outperforms standard LoRA and several advanced PEFT variants across ranking metrics at the same low-rank budget. These results suggest that principled geometric PEFT can improve the quality of social-science concept retrieval in web infrastructures without increasing the number of trainable parameters, which is important for energy- and resource-constrained Web4Good deployments.
\end{itemize}

\section{Related Work}
\label{sec:related_work}

Our work is situated at the intersection of parameter-efficient fine-tuning (PEFT) and geometric deep learning. We first review LoRA and its architectural variants, framing them as addressing the symptoms of underlying geometric issues. We then analyze the emerging use of orthogonality in PEFT, arguing that it has been used as a heuristic patch rather than a foundational principle. Finally, we position OrthoGeoLoRA as a principled geometric remedy that resolves these foundational issues.

\subsection{LoRA and its Architectural Variants}
The challenge of adapting large pre-trained models \citep{devlin-etal-2019-bert, radford2019language, NEURIPS2020_1457c0d6} has made PEFT the dominant paradigm for transfer learning \citep{lialin2024scalingscaleupguide, han2024parameterefficientfinetuninglargemodels}. Among numerous techniques like adapter-based \citep{houlsby2019parameterefficienttransferlearningnlp} and prompt-based tuning \citep{lester-etal-2021-power}, Low-Rank Adaptation (LoRA) \citep{DBLP:journals/corr/abs-2106-09685} has become the de facto standard. It operates on the hypothesis that weight updates reside in a low "intrinsic rank" subspace \citep{aghajanyan-etal-2021-intrinsic}, injecting a trainable update $\Delta W = BA^\top$ while freezing the base weights.

While empirically successful, LoRA's simple parameterization, $\Delta W = BA^\top$, conceals the geometric deficiencies we identify: gauge freedom, scale ambiguity, and a propensity for rank collapse. Much of the follow-up work has focused on architectural modifications that address the *symptoms* of these issues, rather than their root cause.
\begin{itemize}
    \item \textbf{Decoupling Magnitude and Direction:} Weight-Decomposed Low-Rank Adaptation (DoRA) \citep{liu2024doraweightdecomposedlowrankadaptation} decomposes the pre-trained weight into magnitude and direction, applying the LoRA update only to the direction. This directly addresses a symptom of the \emph{scale ambiguity} we identified, but it provides a heuristic fix without altering the core $BA^\top$ parameterization that creates the ambiguity.
    \item \textbf{Efficient Rank Allocation:} AdaLoRA \citep{zhang2023adaloraadaptivebudgetallocation} allocates the parameter budget by parameterizing the update in an SVD-like form and dynamically pruning singular values. This is a form of structured pruning that tackles the symptom of wasted capacity caused by \emph{rank collapse}. In contrast, our approach aims to prevent rank collapse by construction.
    \item \textbf{Expanding Expressivity:} Methods like LoHa \citep{hyeonwoo2023fedparalowrankhadamardproduct} and LoKr \citep{edalati2022kronaparameterefficienttuning} use Hadamard and Kronecker products, respectively, to achieve a higher effective rank with fewer parameters. Their goal of enhancing model \textit{capacity} is orthogonal to ours, which is to improve the \textit{geometric quality} and \textit{optimization stability} of the parameter space itself.
\end{itemize}

\subsection{Orthogonality as a Heuristic in LoRA Variants}
\label{sec:orthogonality_as_heuristic}
A distinct line of work has incorporated orthogonality into LoRA. However, a close analysis reveals that these methods use orthogonality as a heuristic tool to achieve specific, often external goals, leaving the internal geometry of the LoRA update matrix $\Delta W = BA^\top$ untouched. We categorize them as follows:

\paragraph{Orthogonality Relative to Pre-trained Weights.} One family of methods enforces orthogonality with respect to the frozen weights $W_0$. Orthogonal-initialization approaches \citep{buyukakyuz2024olora} use factors of $W_0$ to initialize $A$ and $B$, while projection-based methods \citep{xiong2025oplora} project the update onto the orthogonal complement of $W_0$'s dominant singular directions. These strategies constrain \emph{where} the LoRA update lives relative to the base model—primarily to preserve pre-trained knowledge—but not \emph{how} the update itself is parameterized. The internal gauge freedom, scale ambiguity, and rank collapse of the $BA^\top$ structure persist throughout training.

\paragraph{Orthogonality Across Multiple Tasks.} In multi-task and continual learning, methods like O-LoRA \citep{wang2023orthogonal} and N-LoRA \citep{guo2025nlora} encourage the low-rank subspaces for different tasks to be mutually orthogonal. This mitigates catastrophic forgetting by preventing task-specific updates from interfering with each other. While effective, these approaches treat each LoRA adapter as a black box; orthogonality is enforced \emph{between} adapters, but the geometric deficiencies within each task-specific $\Delta W_t = B_t A_t^\top$ remain unsolved.

In summary, prior work uses orthogonality as a "patch" to solve external problems like knowledge preservation or task interference. They do not re-examine the fundamental geometric soundness of the LoRA parameterization itself.

\section{The Geometric Deficiencies of LoRA}

To understand what these orthogonality-based patches are implicitly compensating for, we now examine the geometry of the vanilla LoRA parameterization itself. The standard LoRA update is parameterized as $\Delta W = BA^T$, where $B \in \mathbb{R}^{d_{out} \times r}$ and $A \in \mathbb{R}^{d_{in} \times r}$. While parameter-efficient, this formulation suffers from several geometric flaws. 

\subsection{Gauge Freedom: Redundancy in Parameter Space}
For any invertible matrix $M \in \mathbb{R}^{r \times r}$, the parameterization is invariant under the transformation $B' = BM$ and $A' = A(M^{-1})^T$, since $B'(A')^T = BA^T$. This creates continuous subspaces of equivalent solutions in the parameter space, forming flat valleys in the loss landscape that can impede optimizer convergence.

\subsection{Scale Ambiguity: Confounded Direction and Magnitude}
The scale of the update can be arbitrarily distributed between factors, as $BA^T = (cB)(\frac{1}{c}A)^T$ for any scalar $c \neq 0$. This ambiguity complicates the interpretation of learning rates and the application of regularization like weight decay.

\subsection{Rank Collapse: Degeneracy of Learned Features}
Without constraints, Euclidean gradient descent often encourages the column vectors of $A$ and $B$ to become collinear to reinforce dominant feature directions. This leads to the effective rank of $\Delta W$ being lower than the configured rank $r$, wasting model capacity.

\subsection{The Ideal Structure: SVD as a Blueprint}
To understand how one might mitigate these geometric pathologies, it is useful to consider an idealized factorization of $\Delta W$. The Eckart-Young-Mirsky theorem states that the best rank-$r$ approximation of a matrix is given by its truncated Singular Value Decomposition (SVD), $\Delta W \approx U_r \Sigma_r V_r^T$. This decomposition is ideal because the orthogonality of $U_r$ and $V_r$ eliminates redundancy and rank collapse, while the diagonal matrix $\Sigma_r$ cleanly decouples feature direction from magnitude. This provides the blueprint for our proposed method.

\subsection{Our Contribution: A Geometric Reparameterization Perspective}
OrthoGeoLoRA's design is grounded in the principles of geometric deep learning, shifting the perspective from heuristic patches to a foundational repair of LoRA's parameter space.

\paragraph{The Geometric Ideal.} Enforcing orthogonality on neural network weights is known to stabilize training, reduce feature redundancy, and improve the loss landscape's conditioning \citep{bansal2018gainorthogonalityregularizationstraining, huang2020controllableorthogonalizationtrainingdnns}. The Eckart--Young--Mirsky theorem establishes that the optimal rank-$r$ approximation to a matrix is its truncated SVD, $\Delta W^\star = U_r \Sigma_r V_r^\top$. This structure, with its orthonormal factors ($U_r, V_r$) and decoupled scaling ($\Sigma_r$), is the ideal geometric form that inherently avoids rank collapse, scale ambiguity, and gauge freedom (up to trivial permutations).

\paragraph{Optimization on the Stiefel Manifold via Reparameterization.} The set of matrices with orthonormal columns forms the Stiefel manifold, $St(d, r)$. Directly optimizing over this curved space would require specialized Riemannian optimizers \citep{10.5555/1557548}, hindering broad adoption. We bypass this by employing \emph{geometric reparameterization} \citep{Casado2019CheapOC}. We maintain unconstrained parameters in Euclidean space, which are passed through a differentiable mapping (e.g., QR decomposition) to produce on-manifold parameters at every forward pass. This makes the orthogonality constraint transparent to standard optimizers like Adam, which retain their internal states.

\paragraph{Positioning OrthoGeoLoRA.} By parameterizing the LoRA update as $\Delta W = B \Sigma A^\top$ with $A \in \mathrm{St}(d_{\mathrm{in}}, r)$ and $B \in \mathrm{St}(d_{\mathrm{out}}, r)$ via this reparameterization technique, OrthoGeoLoRA becomes the first method, to our knowledge, to reform LoRA's internal geometry to match the optimal SVD structure. This principled approach directly resolves the geometric flaws of the standard $BA^\top$ formulation by construction, rather than merely mitigating their symptoms. Our work thus forges a novel link between the practical art of parameter-efficient fine-tuning and the principled science of geometric optimization.

\section{OrthoGeoLoRA: Fine-Tuning on the Stiefel Manifold}

\subsection{Parameterizing with the Stiefel Manifold}
Inspired by the SVD, we reformulate the LoRA update as:
\begin{equation}
\Delta W \;=\; B\,\Sigma\,A^{\top},
\end{equation}
where $\Sigma = \mathrm{diag}(\sigma_1,\dots,\sigma_r)$ is a learnable diagonal matrix of singular values. The factor matrices $A$ and $B$ are constrained to lie on the Stiefel manifold, the space of $d \times r$ matrices with orthonormal columns:
\begin{equation}
\mathrm{St}(d,r) \;=\; \{ X \in \mathbb{R}^{d \times r} : X^{\top} X = I_r \}.
\end{equation}
In particular, we enforce
\begin{equation}
A \in \mathrm{St}(d_{\text{in}},r),\qquad
B \in \mathrm{St}(d_{\text{out}},r).
\end{equation}

\subsection{Optimization via Geometric Reparameterization}
\label{subsec:geom-reparam}
Directly optimizing orthogonality-constrained factors on the Stiefel manifold
(e.g., $A\!\in\!\mathrm{St}(d_{\text{in}},r)$, $B\!\in\!\mathrm{St}(d_{\text{out}},r)$)
requires Riemannian optimizers. However, naively projecting after each Euclidean
update disrupts optimizer states (e.g., momentum in Adam). To avoid this, we
employ \emph{geometric reparameterization} (a.k.a.\ trivialization): we optimize
unconstrained Euclidean parameters and map them \emph{deterministically and
differentiably} to valid manifold elements at every forward pass.

\paragraph{Parameterization.}
Rather than optimizing $(A,B,\Sigma)$ directly, we maintain \emph{unconstrained}
parameters $(\widehat A,\widehat B,s)$ in Euclidean spaces and define
\begin{equation}
A \;=\; \mathrm{Orth}(\widehat A),\quad
B \;=\; \mathrm{Orth}(\widehat B),\quad
\end{equation}
\begin{equation}
\Sigma \;=\; \mathrm{diag}\!\big(\mathrm{softplus}(s)+\varepsilon\big),
\end{equation}
where $\mathrm{Orth}(\cdot)$ produces column-orthonormal factors (Householder) and $\varepsilon{>}0$ ensures numerical stability and
non-negativity of the diagonal entries.

\paragraph{Initialization.}
We initialize
$\widehat A\!\sim\!\mathcal N(0,1)$,
$\widehat B\!\sim\!\mathcal N(0,1)$,
and $s\!=\!\mathbf 0$ (optionally with a small global scale for
$\Sigma$ such as $\alpha/r$). This yields a near-zero update at start,
i.e., $\Delta W\!\approx\!0$, providing a safe warm start.

\paragraph{Training loop (drop-in).}
Let $x\!\in\!\mathbb{R}^{d_{\text{in}}}$. Each step proceeds as:
\begin{enumerate}
  \item \textbf{Forward:} Build $A,B,\Sigma$ from $(\widehat A,\widehat B,s)$
  as above (guaranteeing $A^{\top}\!A=I$, $B^{\top}\!B=I$, $\Sigma\succeq 0$),
  then compute $\Delta h=B\,\Sigma\,A^{\top}x$ and $h=Wx+\Delta h$.
  \item \textbf{Backward:} Backpropagate gradients through the differentiable
  maps $\,\widehat A\!\mapsto\!A$, $\,\widehat B\!\mapsto\!B$, $\,s\!\mapsto\!\Sigma$,
  i.e., $\nabla_{\widehat A}L$, $\nabla_{\widehat B}L$, $\nabla_s L$ are obtained
  by the chain rule; the Jacobians implicitly encode the manifold geometry.
  \item \textbf{Optimizer step:} Apply Adam/AdamW \emph{in the Euclidean spaces}
  of $(\widehat A,\widehat B,s)$, preserving their internal states (momenta,
  second moments). No special Riemannian optimizer or after-the-fact projection
  is required.
\end{enumerate}

This design cleanly separates \emph{subspace selection} (orthonormal $A,B$) from
\emph{per-direction scaling} (diagonal $\Sigma$), removes gauge/scale
ambiguities, and mitigates rank collapse while remaining a \emph{drop-in}
enhancement to standard training pipelines.

\begin{figure*}[htbp]
  \centering
  % The figure is a full-page style method diagram you provided.
  \includegraphics[width=\linewidth]{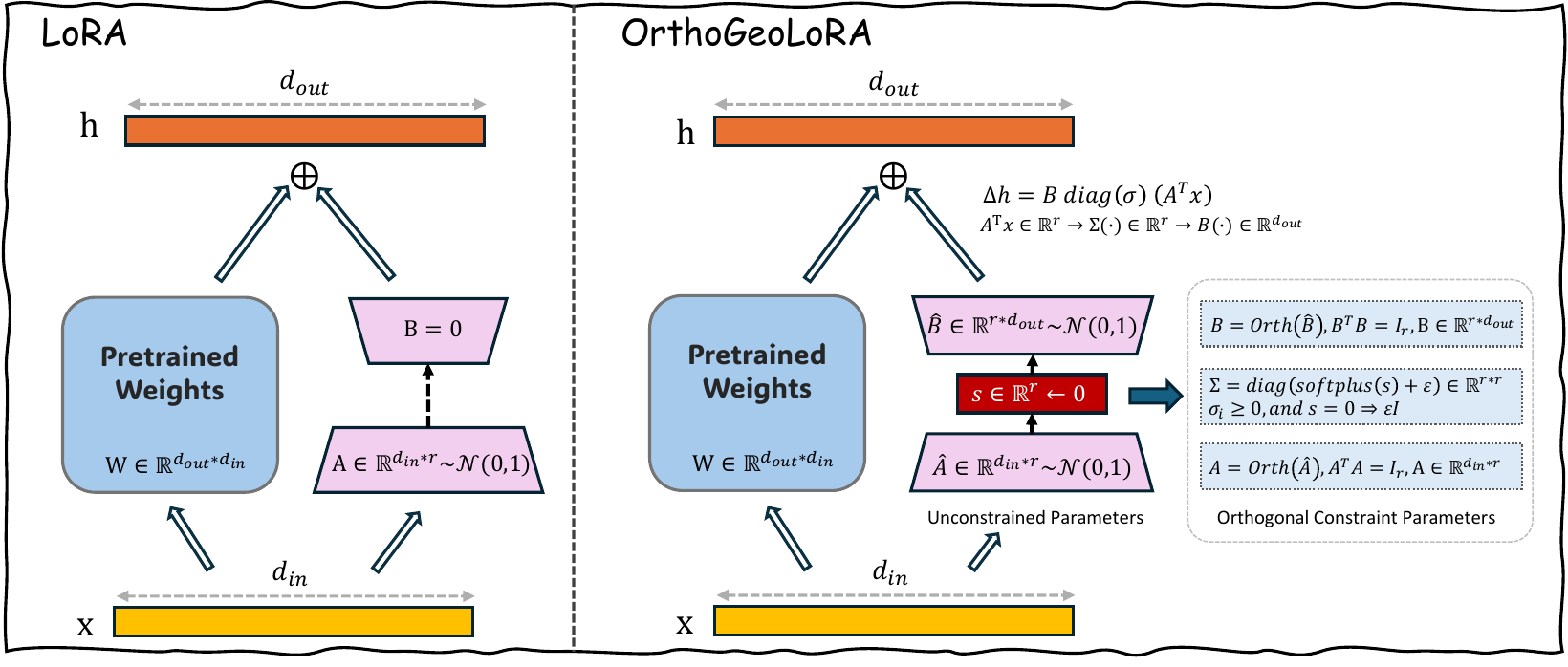} % <-- replace path if needed
  \caption{\textbf{OrthoGeoLoRA overview.}
  Unconstrained Euclidean parameters $(\widehat A,\widehat B,s)$ are mapped
  to orthonormal factors $A,B$ and a non-negative diagonal $\Sigma$, yielding
  $\Delta W=B\,\Sigma\,A^{\top}$. The data flow $x\!\rightarrow\!A^{\top}x
  \!\rightarrow\!\Sigma(\cdot)\!\rightarrow\!B(\cdot)$ is shown together with
  dimensions and constraints; see text for details.}
  \label{fig:overall-geom}
\end{figure*}

\subsection{Practical Implementation}
\label{subsec:implementation}

The core mechanism of OrthoGeoLoRA can be distilled into the procedure outlined in Algorithm~\ref{alg:core_method}. The method modifies a pre-trained linear layer, defined by its weight matrix $W_0 \in \mathbb{R}^{d_{\text{out}} \times d_{\text{in}}}$, by adding a low-rank delta update.

\begin{algorithm}[!htbp]
\caption{The OrthoGeoLoRA Update Mechanism}
\label{alg:core_method}
\begin{algorithmic}[1]
\STATE \textbf{Input:} Pre-trained weight $W_0$, input vector $x \in \mathbb{R}^{d_{\text{in}}}$.
\STATE \textbf{Hyperparameters:} Rank $r$, scaling factor $\alpha$.
\STATE
\STATE \textbf{Learnable Parameters:}
\STATE \quad $\Theta_A \in \mathbb{R}^{d_{\text{in}} \times r}$ \COMMENT{Hidden parameters for A}
\STATE \quad $\Theta_B \in \mathbb{R}^{d_{\text{out}} \times r}$ \COMMENT{Hidden parameters for B}
\STATE \quad $\Sigma \in \mathbb{R}^{r}$ \COMMENT{Singular values}
\STATE
\STATE \textbf{Initialization:}
\STATE \quad Initialize $\Theta_A, \Theta_B$ (e.g., Kaiming uniform).
\STATE \quad $\Sigma \gets \mathbf{0}$.
\STATE
\STATE \textbf{Forward Computation:}
\STATE \quad \COMMENT{1. Apply geometric reparameterization to get orthogonal factors.}
\STATE \quad $A \gets \mathcal{F}_A(\Theta_A)$, where $A \in St(d_{\text{in}}, r)$
\STATE \quad $B \gets \mathcal{F}_B(\Theta_B)$, where $B \in St(d_{\text{out}}, r)$
\STATE
\STATE \quad \COMMENT{2. Compute the delta update $\Delta W(x) = (B \cdot \text{diag}(\Sigma) \cdot A^T)x$.}
\STATE \quad $h \gets xA$ \COMMENT{$h \in \mathbb{R}^{r}$}
\STATE \quad $h \gets h \odot \Sigma$ \COMMENT{$\odot$ is element-wise product}
\STATE \quad $\Delta y \gets hB^T$ \COMMENT{$\Delta y \in \mathbb{R}^{d_{\text{out}}}$}
\STATE
\STATE \quad \COMMENT{3. Combine with the base layer output.}
\STATE \quad $y_0 \gets W_0x$
\STATE \quad $y \gets y_0 + \frac{\alpha}{r} \Delta y$
\STATE
\STATE \quad \textbf{return} $y$
\end{algorithmic}
\end{algorithm}

The key to this implementation is the geometric reparameterization step (Lines 12-13). The learnable parameters that the optimizer (e.g., Adam) sees and updates are the unconstrained matrices $\Theta_A$ and $\Theta_B$, which reside in a standard Euclidean space. The function $\mathcal{F}$ represents a deterministic and differentiable mapping from this Euclidean space to the Stiefel manifold ($St(d, r)$), the space of matrices with orthonormal columns. We implement $\mathcal{F}$ efficiently using the orthogonal parametrization utilities provided by PyTorch, which are typically based on Householder reflections.

This formulation elegantly solves the core optimization challenge:
\begin{itemize}
    \item The orthogonality constraints on $A$ and $B$ are strictly enforced by the very nature of the map $\mathcal{F}$.
    \item The optimizer performs unconstrained updates on $\Theta_A$ and $\Theta_B$, preserving its internal states (e.g., momentum) and requiring no modification to standard training loops.
    \item The direction ($A, B$) and magnitude ($\Sigma$) of the update are mathematically decoupled, leading to a more stable and well-conditioned optimization problem.
\end{itemize}
The application of this layer to a full model follows a standard procedure of identifying target linear modules and replacing them in-place with our adapter, after which only the parameters $\Theta_A, \Theta_B, \Sigma$ are marked as trainable.

\subsection{Complexity and Parameter Efficiency}
\label{sec:complexity}

We briefly compare the parameter counts and computational complexity of OrthoGeoLoRA with standard LoRA and full fine-tuning for a single linear layer. Let the base weight be $W_0 \in \mathbb{R}^{d_{\text{out}} \times d_{\text{in}}}$ and the LoRA rank be $r$.

\paragraph{Trainable parameters.}
In full fine-tuning, all entries of $W_0$ are updated, so the number of trainable parameters for this layer is
\begin{equation}
    P_{\text{full}} = d_{\text{out}} \cdot d_{\text{in}}.
\end{equation}
Standard LoRA introduces a low-rank update $\Delta W = BA^\top$ with $A \in \mathbb{R}^{d_{\text{in}} \times r}$ and $B \in \mathbb{R}^{d_{\text{out}} \times r}$, while freezing $W_0$. The trainable parameters are then
\begin{equation}
    P_{\text{LoRA}} = d_{\text{in}} \cdot r + d_{\text{out}} \cdot r.
\end{equation}
OrthoGeoLoRA adopts an SVD-like form $\Delta W = B \Sigma A^\top$ and uses unconstrained Euclidean parameters $\Theta_A, \Theta_B \in \mathbb{R}^{d \times r}$ and $s \in \mathbb{R}^r$ which are mapped to $A,B \in \mathrm{St}(\cdot, r)$ and $\Sigma = \mathrm{diag}(\mathrm{softplus}(s) + \varepsilon)$. The number of trainable parameters is therefore
\begin{equation}
    P_{\text{OrthoGeoLoRA}} = d_{\text{in}} \cdot r + d_{\text{out}} \cdot r + r.
\end{equation}
Thus, OrthoGeoLoRA matches the parameter budget of standard LoRA up to an additive $\mathcal{O}(r)$ term for the singular values, and both are dramatically more parameter-efficient than full fine-tuning. For a square layer with $d_{\text{in}} = d_{\text{out}} = d$, the ratio between OrthoGeoLoRA and full fine-tuning is approximately
\begin{equation}
    \frac{P_{\text{OrthoGeoLoRA}}}{P_{\text{full}}}
    \approx \frac{2 d r}{d^2}
    = \frac{2r}{d},
\end{equation}
which is typically well below $1\%$ for $r \ll d$.

\paragraph{Forward computation.}
The forward pass of the base linear layer, $y_0 = W_0 x$ for $x \in \mathbb{R}^{d_{\text{in}}}$, costs $\mathcal{O}(d_{\text{out}} d_{\text{in}})$ operations. Standard LoRA adds a low-rank update $\Delta y = BA^\top x$, which can be computed as
\begin{align}
    u &= A^\top x &&\in \mathbb{R}^r, \\
    \Delta y &= B u &&\in \mathbb{R}^{d_{\text{out}}},
\end{align}
with an additional cost of $\mathcal{O}(d_{\text{in}} r + d_{\text{out}} r)$. OrthoGeoLoRA uses
\begin{align}
    u &= A^\top x &&\in \mathbb{R}^r, \\
    \tilde{u} &= \Sigma u &&\in \mathbb{R}^r, \\
    \Delta y &= B \tilde{u} &&\in \mathbb{R}^{d_{\text{out}}},
\end{align}
which adds an extra $\mathcal{O}(r)$ term for the diagonal scaling. The forward-time complexity is therefore the same asymptotic order as LoRA:
\begin{equation}
    \mathcal{O}\big(d_{\text{out}} d_{\text{in}}\big)
    \;+\; \mathcal{O}\big( (d_{\text{in}} + d_{\text{out}}) r \big).
\end{equation}
Since $r \ll d_{\text{in}}, d_{\text{out}}$ in typical PEFT setups, the dominant cost remains the base matrix--vector product $W_0 x$.

\paragraph{Overhead of orthogonalization.}
The main additional cost of OrthoGeoLoRA arises from the orthogonalization mapping $\Theta_A, \Theta_B \mapsto A,B$ at each training step. For QR- or Householder-based implementations, this costs $\mathcal{O}(d_{\text{in}} r^2 + d_{\text{out}} r^2)$ per layer and parameter update. For small ranks (e.g., $r=8$ in our experiments) and transformer dimensions in the hundreds, this term is dominated by the $\mathcal{O}(d_{\text{out}} d_{\text{in}})$ cost of the base model. Importantly, orthogonalization is only required during training; at inference time, the composed update $\Delta W = B \Sigma A^\top$ can be precomputed and folded into $W_0$, so the runtime cost is identical to that of standard LoRA.

\paragraph{Implications for Web4Good.}
From a Web4Good perspective, these observations imply that OrthoGeoLoRA preserves the parameter- and memory-efficiency benefits of LoRA while providing better-structured adaptations. Smaller parameter sets reduce optimizer state and checkpoint size, and avoiding full fine-tuning keeps the marginal training cost of adapting models manageable for resource-constrained institutions, even when orthogonalization is taken into account.

\section{Experimental Setup}

\subsection{Task and Dataset}
\label{subsec:task_dataset}

We evaluate OrthoGeoLoRA on a hierarchical concept retrieval task that is directly motivated by social science web infrastructures. Concretely, we consider the problem of retrieving controlled-vocabulary concepts from the European Language Social Science Thesaurus (ELSST) given free-text descriptions. ELSST is a curated, multilingual ontology used by data archives and digital libraries to index social science studies and resources. In practice, web interfaces built on top of such thesauri must map heterogeneous user queries---ranging from technical abstracts to policy-oriented descriptions---onto standardized concepts to support search, browsing, and cross-lingual interoperability.

Formally, each instance consists of a short textual description and a fixed candidate set of ELSST concepts. The model encodes the description and the candidate labels and is evaluated on its ability to rank the correct concept highly. This framing reflects realistic usage scenarios of social science web portals, where users seldom know the exact thesaurus term but nevertheless need to discover the right concept. It also emphasizes the need for structured, ontology-aware representations, making it a natural testbed for our geometric PEFT method.

\paragraph{ELSST-based Synthetic Dataset.} We constructed a core English training set using concepts from ELSST. To move beyond simple keyword matching and teach the model robust semantic representations, we generated diverse descriptive texts for each concept using a powerful language model (DeepSeek-V3). The generation process employed a variety of prompts to elicit nuanced expressions, incorporating different contexts (e.g., academic, applied), perspectives (e.g., researcher, policymaker), and styles. This synthetic dataset, structured according to the ELSST hierarchy, was used for all model training.

\paragraph{Data quality assurance.}
The practical usefulness of our synthetic dataset depends critically on its linguistic and conceptual quality. To obtain an initial check, we asked two domain experts to rate a random sample of 40 concept--text pairs on (i) conceptual accuracy (1--5 scale) and (ii) linguistic fluency (1--3 scale). The descriptive statistics and inter-annotator agreement, reported in Appendix~\ref{sec:appendix_dataset_validation} (Table~\ref{tab:human_validation_results}), indicate that the synthetic descriptions are generally accurate and fluent enough to support our experiments, while still leaving room for future refinement.

\subsection{Baselines}
We compare OrthoGeoLoRA against several strong baselines:
\begin{itemize}

    \item \textbf{Zero-Shot Base Model:}: The original untrained base model (serving as a reference).
    \item \textbf{LoRA}: The standard low-rank adaptation method.
    \item \textbf{LoRA Variants}: We include several advanced variants: \textbf{AdaLoRA}, \textbf{DoRA}, \textbf{LoHa}, and \textbf{LoKr} to ensure a comprehensive comparison.

\end{itemize}

\subsection{Implementation Details}
All experiments use \texttt{multilingual-e5-small} as the base model. For all PEFT methods, we set the rank $r=8$. We use the AdamW optimizer with a learning rate of $1e-4$ and a batch size of $128$.

\subsection{Evaluation Metrics}
We evaluate retrieval performance using Mean Reciprocal Rank (MRR), Recall@k, and Normalized Discounted Cumulative Gain (NDCG@k) for k=\{1, 3\}.

\section{Results and Analysis}

\subsection{Main Performance}
Table~\ref{tab:main_results} presents the main results on our ELSST concept retrieval benchmark. OrthoGeoLoRA consistently outperforms all other methods across all evaluation metrics. In particular, it improves Recall@3 by +4.1 points and NDCG@3 by +2.8 points over the next best PEFT variant, indicating that it more reliably places the correct concept among the top-ranked candidates on this benchmark. It is also worth noting that some complex LoRA variants, such as AdaLoRA, underperform even standard LoRA on this structured task, suggesting that, at least for this structured retrieval task, principled geometric constraints can be more beneficial than adding further architectural complexity.

% Main Results Table
\begin{table}[!htbp]
\centering
\caption{Main results on the ELSST concept retrieval task. Best scores are in \textbf{bold}.}
\label{tab:main_results}
\resizebox{\columnwidth}{!}{%
\begin{tabular}{lccccc}
\toprule
\textbf{Method} & \textbf{MRR} & \textbf{Recall@1} & \textbf{Recall@3} & \textbf{NDCG@1} & \textbf{NDCG@3} \\
\midrule
\textbf{Base} & 0.954 & 0.304 & 0.831 & 0.939 & 0.889 \\
\midrule
LoRA         & 0.973 & 0.315 & 0.898 & 0.963 & 0.936 \\
AdaLoRA      & 0.948 & 0.299 & 0.833 & 0.923 & 0.882 \\
DoRA         & 0.974 & 0.316 & 0.894 & 0.965 & 0.935 \\
LoHa         & 0.961 & 0.308 & 0.846 & 0.947 & 0.897 \\
LoKr         & 0.955 & 0.304 & 0.850 & 0.937 & 0.898 \\
\midrule
\textbf{OrthoGeoLoRA} & \textbf{0.983} & \textbf{0.321} & \textbf{0.939} & \textbf{0.978} & \textbf{0.964} \\
\bottomrule
\end{tabular}%
}
\end{table}

\subsection{Analysis: Why Does Orthogonality Help?}
The superior performance of OrthoGeoLoRA can be attributed directly to its geometric constraints. By enforcing orthogonality, our method prevents rank collapse, ensuring that the model utilizes the full capacity of the configured rank $r$ to learn distinct, non-redundant conceptual features. The explicit decoupling of direction ($A, B$) and magnitude ($\Sigma$) provides a cleaner, more stable optimization landscape.

\subsection{Empirical Validation of Geometric Benefits}
\label{subsec:empirical_validation}

To provide direct empirical evidence for these claims, we conduct a series of detailed analyses visualizing the optimization process and the properties of the learned representations.

\begin{figure}[h!]
    \centering
    \includegraphics[width=0.9\columnwidth]{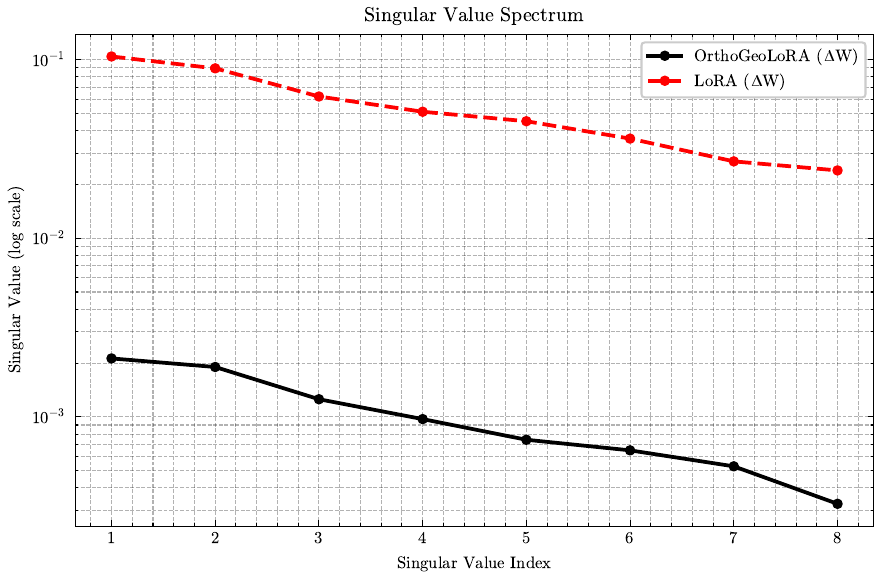} % Make sure the path is correct
    \caption{Singular value spectrum of the learned update matrix ($r=8$). LoRA's sharp decay is a clear indicator of rank collapse, while OrthoGeoLoRA maintains a healthier spectrum, demonstrating full utilization of its allocated rank.}
    \label{fig:singular_values}
\end{figure}

\paragraph{Solving Rank Collapse.}
Figure~\ref{fig:singular_values} offers a direct visualization of how OrthoGeoLoRA solves the rank collapse problem. We plot the singular value spectrum of the learned update matrix for both LoRA (obtained by performing SVD on the composed $BA^T$ matrix) and OrthoGeoLoRA (the learned $\Sigma$ values). Standard LoRA exhibits a steep drop-off, a classic symptom of rank collapse where only a few dominant singular values are learned, wasting the allocated parametric capacity. In stark contrast, OrthoGeoLoRA maintains a significantly flatter, healthier spectrum, indicating that all $r$ dimensions are being effectively utilized to capture distinct features. This pattern is consistent with the intended effect of our geometric constraints and suggests that they help prevent
the degeneracy of learned representations.

\begin{figure}[h!]
    \centering
    \includegraphics[width=0.9\columnwidth]{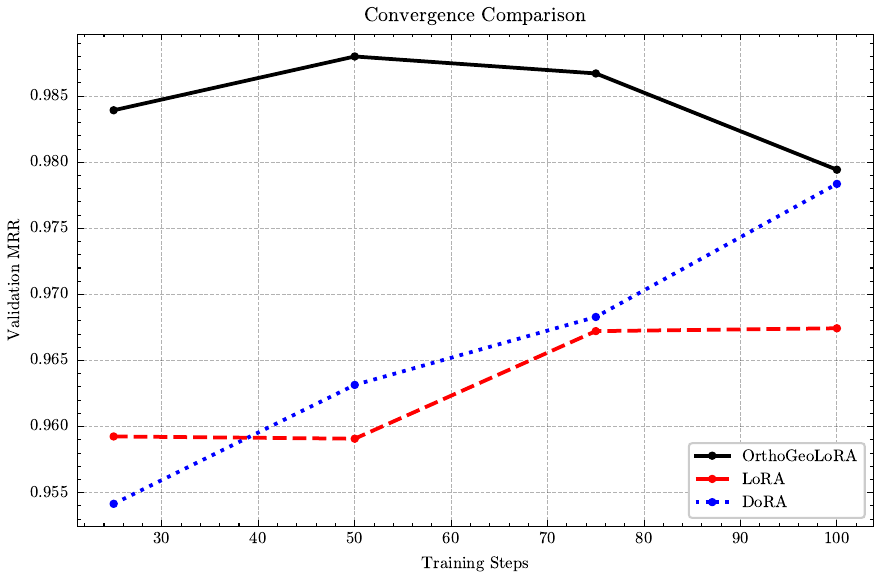} % Make sure the path is correct
    \caption{Validation MRR over training steps. OrthoGeoLoRA exhibits faster convergence and achieves a higher peak performance, indicating a more efficient optimization landscape.}
    \label{fig:convergence}
\end{figure}

\paragraph{Improved Optimization Dynamics.}
Our claim of a "cleaner, more stable optimization landscape" is empirically supported by the convergence analysis in Figure~\ref{fig:convergence}. OrthoGeoLoRA not only converges to a significantly higher final performance (MRR) but also does so more rapidly than both LoRA and DoRA. Its learning curve demonstrates a faster ascent and greater stability from the early stages of training. This improved optimization behaviour is consistent with the hypothesis that reducing redundant parameterizations and ambiguities in the update space leads to a better-conditioned problem that is more amenable to standard gradient-based optimizers.

\begin{figure}[h!]
    \centering
    \includegraphics[width=0.9\columnwidth]{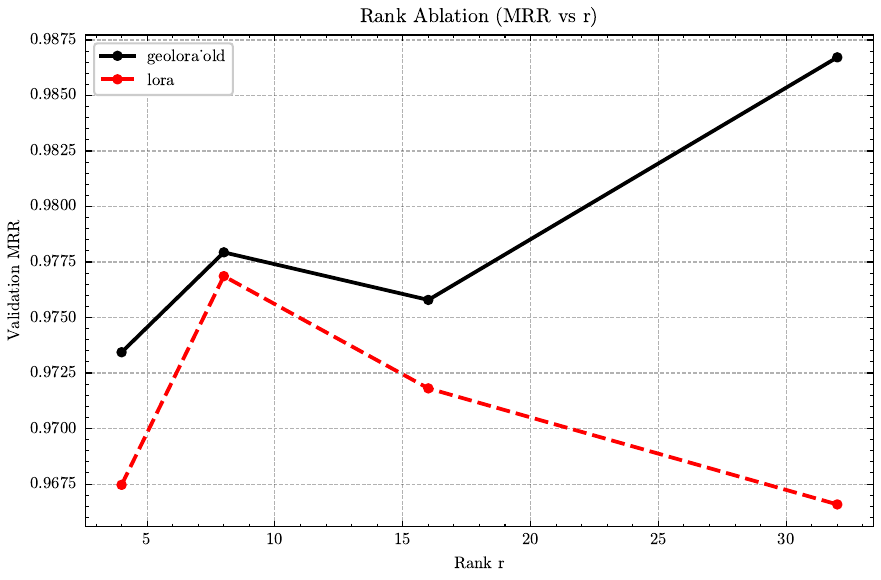} % Make sure the path is correct
    \caption{Performance (MRR) as a function of rank $r$. OrthoGeoLoRA consistently outperforms LoRA and scales more effectively with increased parametric capacity.}
    \label{fig:rank_ablation}
\end{figure}

\paragraph{Superior Parameter Efficiency.}
Finally, we analyze the impact of the rank hyperparameter $r$ on performance in Figure~\ref{fig:rank_ablation}. The results highlight two critical findings. First, OrthoGeoLoRA consistently outperforms LoRA across all tested ranks, demonstrating the robustness of its benefits across different parameter budgets. Second, and more strikingly, LoRA's performance peaks at $r=8$ and degrades with a higher rank, suggesting its unstable parameterization prevents it from effectively utilizing a larger capacity. Conversely, OrthoGeoLoRA shows a healthy scaling trend, confirming its ability to leverage additional parameters effectively. Notably, an $r=8$ OrthoGeoLoRA surpasses even a higher-capacity $r=16$ LoRA, underscoring its superior parameter efficiency.

\section{Societal and Ethical Considerations}
\label{sec:societal_ethical}

\paragraph{Social science web infrastructures and public good.}
Our evaluation task is grounded in ELSST, a controlled vocabulary used to organize social science resources in web-based repositories. Improving concept retrieval in this setting can support more effective discovery of studies on topics such as inequality, migration, and public health by researchers, policymakers, and NGOs. In this sense, OrthoGeoLoRA contributes a technical building block for web infrastructures that serve the public good, rather than optimizing for commercial recommendation alone.

\paragraph{Compute, accessibility, and sustainability.}
Because OrthoGeoLoRA operates at the same low-rank budget as standard LoRA and uses standard optimizers, it can be deployed wherever LoRA is currently feasible. Our results therefore indicate that better structured representations can be obtained \emph{without} increasing the number of trainable parameters. While we do not directly measure energy usage, parameter-efficient and geometry-aware PEFT methods of this kind can play a role in reducing the marginal compute cost of adapting foundation models, which is especially important for smaller institutions and regions with limited computational resources. We see this work as complementary to more explicit energy-accounting approaches in sustainable AI.

\paragraph{Data, bias, and synthetic text.}
Our dataset construction pipeline relies on a large language model to generate synthetic descriptions conditioned on ELSST concepts and definitions. Although ELSST itself is a curated ontology, the generated texts may still reflect biases present in the language model's training data, for example in how social categories or policy issues are described. We partially mitigate this by focusing on implicit, concept-level descriptions and by validating a sample of the data with domain experts, but we do not claim to remove all forms of bias. Future work could combine OrthoGeoLoRA with explicit fairness constraints or contrastive data augmentation to better control for such effects, particularly when deploying similar systems in high-stakes decision-support scenarios.

\section{Conclusion}
\label{sec:conclusion}

In this paper, we began by identifying fundamental geometric deficiencies in Low-Rank Adaptation (LoRA), the dominant paradigm for parameter-efficient fine-tuning. We argued that its standard parameterization suffers from gauge freedom, scale ambiguity, and a propensity for rank collapse, which collectively create suboptimal optimization landscapes and limit the model's ability to learn structured representations.

To address these issues, we introduced OrthoGeoLoRA, a principled reformulation of low-rank adaptation inspired by the ideal structure of Singular Value Decomposition. By constraining the low-rank factor matrices to the Stiefel manifold, OrthoGeoLoRA enforces orthogonality, thereby eliminating parameter redundancy and maximizing expressive capacity. Critically, we achieved this through a geometric reparameterization technique, ensuring our method remains a "drop-in" replacement for LoRA, fully compatible with standard optimizers like Adam.

Our empirical evaluation on a novel and challenging benchmark for hierarchical concept retrieval, built upon the ELSST thesaurus, provided strong validation for our approach. The results demonstrated that OrthoGeoLoRA significantly outperforms not only standard LoRA but also a suite of advanced PEFT variants across all metrics. These findings support our central thesis that enforcing a principled geometric structure on the parameter space can lead to more effective and robust fine-tuning, particularly for tasks requiring the mastery of complex, structured
knowledge.

Our work highlights the importance of considering the underlying geometry of parameterization in PEFT and opens up new avenues for developing more structured and efficient model adaptation techniques. Future work could explore other manifold constraints tailored for different inductive biases or apply this geometrically-aware tuning approach to other knowledge-intensive domains, such as legal text analysis or semantic code understanding. Ultimately, OrthoGeoLoRA represents a step towards a new class of PEFT methods grounded, promising a more robust foundation for adapting large models to the structured world they aim to represent.

%\begin{acks}
%To Robert, for the bagels and explaining CMYK and color spaces.
%\end{acks}

\bibliographystyle{ACM-Reference-Format}
\bibliography{sample-base}

\appendix

\appendix

\section{Mathematical Background on Geometric Reparameterization}
\label{sec:appendix_math}

In the main text, we describe geometric reparameterization as a mapping $\mathcal{F}: \mathcal{E} \rightarrow \mathcal{M}$ from a Euclidean space $\mathcal{E}$ to a manifold $\mathcal{M}$. While the specific implementation in PyTorch is based on Householder reflections, the principle can be elegantly illustrated with the \textbf{Cayley transform}, a classical map that connects skew-symmetric matrices to orthogonal matrices.

A real matrix $X$ is skew-symmetric if $X^T = -X$. The set of all $d \times d$ skew-symmetric matrices forms a Euclidean vector space. For a skew-symmetric matrix $X$, the Cayley transform defines a corresponding orthogonal matrix $Q$ as:
\begin{equation}
    Q = (I - X)(I + X)^{-1}
\end{equation}
where $I$ is the identity matrix. This mapping is well-defined as long as $(I+X)$ is invertible.

In the context of OrthoGeoLoRA, the hidden, unconstrained parameter $\Theta$ that the optimizer sees would be a matrix whose columns are used to construct a skew-symmetric matrix $X$. The forward pass would compute the orthogonal matrix $A = Q$ using the Cayley transform. The gradients $\nabla_A L$ would then be backpropagated to $\nabla_\Theta L$ through this transform via the chain rule. This ensures that any standard Euclidean update on $\Theta$ results in a valid movement on the Stiefel manifold of $A$, thus preserving the orthogonality constraint without requiring a custom optimizer.

\section{Dataset Construction and Validation Details}
\label{sec:appendix_dataset}

This section provides a detailed account of our synthetic dataset generation pipeline and the rigorous human validation methodology used to ensure its quality.

\subsection{Synthetic Data Generation Pipeline}
\label{subsec:appendix_generation}

\paragraph{Objective.}
The primary goal of the generation pipeline was to create varied and context-rich descriptions for each concept from the ELSST. A crucial design principle was to generate \textbf{implicit references}—texts that describe a concept's essence without explicitly naming it. This approach compels the model to learn deep semantic representations rather than relying on superficial keyword matching, creating a more challenging and realistic evaluation benchmark.

\paragraph{Generation Strategy.}
We employed a highly structured, template-based prompting strategy defined in a \texttt{YAML} configuration file to guide a large language model (DeepSeek-V3). To ensure a rich diversity of expression, we systematically varied the prompts across several dimensions:

\begin{itemize}
    \item \textbf{Roles and Perspectives:} We defined 16 distinct roles, categorized into \textit{Professional Sociology Roles} and \textit{Social Practice Roles} (see Table~\ref{tab:appendix_prompt_roles}). Each role was associated with specific analytical focus areas (e.g., "power relations" for a critical theorist) and linguistic styles (e.g., "technical," "practical," "accessible").

    \item \textbf{Templates and Sampling:} A mix of role-specific templates (sampled with 70\% probability) and common, general-purpose templates (30\% probability) were used to structure the final prompt. This combination ensured both role-consistent outputs and broader linguistic variety.

    \item \textbf{Strict Implicitness Instruction:} A core instruction block was prepended to every prompt, explicitly forbidding the LLM from mentioning the target concept name or using obvious referring terms (e.g., "this concept," "the aforementioned theory").
\end{itemize}

The pipeline operated by randomly selecting a unique role and a template for each concept. The final prompt, an example of which is shown in Figure~\ref{fig:appendix_prompt_example}, combined the system-level role assignment with the user-level task description. This process was repeated to generate up to 24 unique descriptions per concept.

\begin{table}[h!]
\centering
\caption{The 16 roles utilized in the prompt generation strategy to ensure diverse sociological and practical perspectives.}
\label{tab:appendix_prompt_roles}
\footnotesize
\begin{tabular}{@{}ll@{}}
\toprule
\textbf{Professional Sociology Roles} & \textbf{Social Practice Roles} \\
\midrule
Theoretical Sociologist & Social Worker \\
Empirical Social Researcher & Social Studies Educator \\
Applied Sociologist & Community Organizer \\
Cultural Sociologist & Social Affairs Journalist \\
Social Policy Specialist & Social Impact Consultant \\
Critical Social Theorist & Sociology Student \\
 & Public Administrator \\
 & Healthcare Professional \\
 & NGO Practitioner \\
 & Engaged Citizen \\
\bottomrule
\end{tabular}
\end{table}

\begin{figure}[h!]
    \centering
    \fbox{\begin{minipage}{0.9\linewidth}
    \small
    \textbf{System Prompt:} \\
    You are a critical social theorist. Your task is to analyze a sociological concept without explicitly naming it. \\
    \hrulefill \\
    \textbf{User Prompt (Example):} \\
    Based on the concept "Social Stratification" and its definition "The classification of persons into groups based on shared socio-economic conditions; a relational set of inequalities...", generate an analysis text. \\
    \\
    \textbf{Important requirements:} \\
    1. DO NOT mention the concept name "Social Stratification" directly in your response. \\
    2. DO NOT use obvious referring terms like "this concept", "this theory", etc. \\
    3. Write your description so readers can understand what concept you're discussing without naming it. \\
    4. Use plain text format, not Markdown. \\
    5. Ensure you analyze from the perspective of your assigned role. \\
    \\
    \textbf{Task:} As a critical social theorist, examine relationships to power structures and social inequality. Analyze power relations, social inequalities, systemic critique, transformative potential using critical and transformative analysis, citing critical analyses and social critiques to deepen understanding.
    \end{minipage}}
    \caption{An example of a fully constructed prompt for the concept "Social Stratification" under the role of "Critical Social Theorist." The prompt combines role assignment, strict constraints on implicit expression, and a specific analytical focus.}
    \label{fig:appendix_prompt_example}
\end{figure}

\subsection{Human Validation Methodology}
\label{sec:appendix_dataset_validation}

To assess the reliability and quality of our synthetic dataset, we conducted a structured human evaluation study with domain experts. The evaluation focused on the quality of the ``self'' relation examples, which are critical as they represent varied expressions of a core concept.

\paragraph{Objective and scope.}
We randomly sampled 40 concept--text pairs from the generated dataset. These pairs were evaluated by two independent annotators with graduate-level expertise in sociology. For each pair, annotators were provided with the concept name and its official ELSST definition as ground truth.

\paragraph{Evaluation criteria.}
Annotators assessed each generated text based on two criteria: \emph{Accuracy} and \emph{Fluency}, using the following scales.

\begin{itemize}
    \item \textbf{Accuracy (1--5 scale):} Measures how accurately and unambiguously the generated text represents the core definition of the concept.
    \begin{itemize}
        \item 5 (Highly Accurate): A clear application or explanation that closely matches the definition.
        \item 4 (Basically Accurate): The core idea is consistent with the definition, with only minor extraneous information or suboptimal phrasing.
        \item 3 (Partially Accurate): The text is related to the concept but misses the core point or shows some conceptual confusion.
        \item 2 (Misleading): The text appears related but ultimately misrepresents the concept's definition.
        \item 1 (Completely Wrong): The text is irrelevant or factually incorrect.
    \end{itemize}
    \item \textbf{Fluency (1--3 scale):} Measures the linguistic quality of the text, independent of its conceptual accuracy.
    \begin{itemize}
        \item 3 (Excellent): Natural, fluent, and grammatically correct.
        \item 2 (Acceptable): Generally fluent, but with minor grammatical errors or awkward phrasing that may suggest machine generation.
        \item 1 (Poor): Incoherent, ungrammatical, or difficult to understand.
    \end{itemize}
\end{itemize}

\paragraph{Results.}
Table~\ref{tab:human_validation_results} summarizes the descriptive statistics and inter-annotator agreement for this study. The mean scores suggest that, on average, the generated texts are conceptually accurate and linguistically fluent enough for use in our experiments, while still leaving room for future refinement.

\begin{table}[t]
\centering
\caption{Summary of human validation results for synthetic data quality (N=40). Inter-annotator agreement (IAA) is measured with Krippendorff's Alpha.}
\label{tab:human_validation_results}
\resizebox{0.8\linewidth}{!}{
\begin{tabular}{lcccc}
\toprule
\textbf{Metric} & \textbf{Mean} & \textbf{Median} & \textbf{Std. Dev.} & \textbf{Scale} \\
\midrule
Accuracy & 3.950 & 4.0 & 0.967 & 1--5 \\
Fluency  & 2.712 & 3.0 & 0.482 & 1--3 \\
\midrule
\multicolumn{5}{l}{\textbf{IAA:} 0.85 (Accuracy), 0.72 (Fluency)} \\
\bottomrule
\end{tabular}
}
\end{table}

Inter-annotator agreement was calculated using Krippendorff's Alpha with an ordinal distance function. This protocol and the resulting statistics provide a transparent basis for judging the linguistic and conceptual quality of the synthetic dataset used in our experiments.

\end{document}